\documentclass[journal]{IEEEtran}

\usepackage[utf8]{inputenc}
\usepackage[numbers,sort&compress]{natbib}

\usepackage[table]{xcolor}
\usepackage{times}
\usepackage{epigraph}
\usepackage{epsfig}
\usepackage{graphicx}
\usepackage{rotating}
\usepackage{amsmath}
\usepackage{amssymb}
\usepackage{booktabs}
\usepackage{float}
\usepackage{placeins}
\usepackage{mathabx}
\usepackage{multirow}
\usepackage{balance}
\usepackage{multirow}
\usepackage{soul}
\usepackage{subfigure}
\usepackage[flushleft]{threeparttable}
\usepackage{textcomp}
\usepackage{enumitem}

\setlist[enumerate]{leftmargin=*}
\setlist[itemize]{leftmargin=*}
\setlist[description]{leftmargin=*}

\begin{document}
    
\title{RECOD Titans at ISIC Challenge 2017}

\author{
    Afonso Menegola$^\dag$\thanks{$^\dag$A. Menegola is the main author of Part 3 and J. Tavares is the main author of Part 1.}, Julia Tavares$^\dag$, Michel Fornaciali, Lin Tzy Li, Sandra Avila, Eduardo Valle$^\ast$
    \vspace{-0.75cm}

    \thanks{
        All authors affiliated to the RECOD Lab., and to the Department of Computer Engineering and Industrial Automation (DCA) of the School of Electrical and Computing Engineering (FEEC), University of Campinas (UNICAMP), Campinas, Brazil.
        
        S. Avila and Lin Tzy Li also affiliated to the Institute of Computing (IC), UNICAMP.
        
        $^\ast$Contact author: dovalle@dca.fee.unicamp.br, mail@eduardovalle.com.
    }
}


\maketitle



\section*{History}
	Our team has worked on melanoma classification since early 2014 \cite{fornaciali2014statistical}, and has employed deep learning with transfer learning for that task since 2015 \cite{carvalho2015}. Recently, the community has started to move from traditional techniques towards deep learning, following the general trend of computer vision \cite{fornaciali2016towards}. Deep learning poses a challenge for medical applications, due to the need of very large training sets. Thus, transfer learning becomes crucial for success in those applications, motivating our paper for ISBI 2017~\cite{menegola2017knowledge}.
	
	Our team participated in Parts 1 and 3 of the ISIC Challenge 2017, described below in that order. Although our team has a long experience with skin-lesion classification (Part 3), this Challenge was the very first time we worked on skin-lesion segmentation (Part 1).
	 
\section{Part 1: Lesion Segmentation}

\vspace{-0.1cm}	
\subsection{Baseline}

    From the start, we based our network on the U-net of Ronneberger et al.~\cite{RonnebergerFB15}, a convolutional network intended for accurate segmentation of biomedical images.
    We used a public implementation inspired on U-net\footnote{https://github.com/jocicmarko/ultrasound-nerve-segmentation.git} as a baseline.

\vspace{-0.1cm}	
\subsection{Data and Framework}          
    For training, we employed the ISIC 2017 Challenge official dataset, with 2,000 dermoscopic images. We made some tentative trials with extra data, but those resulted in worse validation scores reported at the validation leaderboard of the challenge website (\textbf{official validation scores}, hereafter) so we decided to use only the official data.
    
    Segmentation is a very subjective activity: studies show that inter-human agreement for the task is far from perfect. Thus, external data, segmented by subjects or criteria different from ISIC's could indeed lead to worse scores. Another explanation just as good is that we did not train our models long enough to profit from the extra data (see the end of Section~\ref{sec:experiments}).
    
    For the models we coded ourselves, we used Keras\footnote{https://keras.io/}, with a Theano backend\footnote{http://deeplearning.net/software/theano/}. All experiments ran on a single NVIDIA GeForce GTX 1070, in the personal computer of the second author.
    
    The code needed to reproduce our results is at our \textbf{code repository}\footnote{Available soon at: https://github.com/learningtitans/isbi2017-part1}.

\vspace{-0.1cm}	
\subsection{Data Augmentation} 
    We used online image augmentation, with up to 10\% horizontal and vertical shifts, up to 20\% zoom, and up to 270° degrees rotation.
    Images were first resized --- we tried 256$\times$256 and 128$\times$128, ultimately keeping the latter, which was faster and resulted in similar performance. Transforming the images before resizing them was slower and did not improve the results.

\vspace{-0.1cm}	
\subsection{Experiments}
\label{sec:experiments}
    Our first attempt was a model based on the VGG network \cite{DBLP:journals/corr/SimonyanZ14a}. The first part of the model consisted of the VGG-16 layers up to the last convolutional layer, but without the last max-pooling layer and the fully-connected layers.  We initialized the retained layers with the weights of a VGG-16 model pre-trained on the ImageNet dataset. We completed the model following the U-shape recipe, with convolutional and up-sampling layers, but without any fully-connected layer. We  added dropout layers after each of the five layers with copied features. Training began by freezing the original layers from VGG-16, and updating the weights of the other layers for 100 epochs; and then, by unfreezing and updating all weights for another 100 epochs. We used the Dice coefficient as loss function. That first model achieved an official validation score of 0.753.
    
    We got our best results with a network based on the work of Codella et al. \cite{codella2015deep} for segmentation. The main changes we applied include: 
    \begin{enumerate}
        \item Dice coefficient as the loss function.
        \item ADAM as the optimizer.
        \item ReLU activations everywhere except in the last layer, where we used the sigmoid activation.
        \item Ground-truth masks rescaled simply by dividing by 255.
        \item Subtracting the mean RGB value of ImageNet training set from input RGB values.
    \end{enumerate}

    We attempted to reduce the first fully-connected layer from 8192 to 4092 or 2048 neurons but results became worse. The best individual network used 8192 neurons and 0.5 dropout everywhere, reaching an official validation score of 0.783 after training for 220 epochs. We were also able to achieve somewhat similar results by removing the fully-connected network altogether (as is the case in the original U-net paper) and adding batch normalization layers. That achieved an official validation score of 0.774, after training for 220 epochs. Although it performed worse than the previous model, it had the advantage of being faster to train and occupying much less disk space due to the lack of a fully-connected layer, but we ultimately did not use it.
    
    \textbf{Our final submission} used four of our models: two trained with all 2000 samples, without a validation split, for 250 and for 500 epochs respectively; and other two trained and validated with two different 1600/400 splits, for 220 epochs. Those four models, individually, achieved between 0.780 and 0.783 official validation scores. Our final submission averaged the output of those four models achieved a score of 0.793.
    
    Alongside the choices that composed the final submission, there were many dead ends, ideas that seemed good, but did not improve our scores:
    \begin{enumerate}
        \item Adding 3 HSV channels to the 3 RGB channels in the input.
        \item Other loss functions, such as the Jaccard index, binary cross entropy, and mean squared error.
        \item Normalizing the input using the mean of each sample, or the mean and standard deviation of all the training dataset, instead of using the global average of all ImageNet training set pixels.
        \item Training with all images with available masks in the ISIC Archive\footnote{http://isdis.net/isic-project/} for 500 epochs --- maybe because the ground-truth masks are not as carefully done in the archive as a whole as they were done for the challenge subset, but maybe simply because the model needed much more time than 500 epochs to converge.
    \end{enumerate}
    
    While it might have been better to train all models for more epochs, due to time constraints, we chose to try new ideas over training the same models longer.  

    As mentioned, the participation in the challenge was the first contact of our research team with skin-lesion segmentation. We would like to explore the interaction between lesion classification and lesion segmentation, and how those tasks can jointly help each other. 

\section{Part 3: Lesion Classification}

\vspace{-0.1cm}	
\subsection{Strategy}


    We aimed, from start, at a deep learning solution. Our previous experience with the technique taught us that three big bottlenecks would limit performance: amount of training data, depth of the learning model, and availability of computational horsepower. Thus, we started by attempting to secure as much data and computational power as possible, in order to use models as deep as possible. 
    
    After those three big issues are solved, there remains the fine craftsmanship of optimizing the models. From start, our aim was to get the highest possible rank at the challenge. If in~\cite{menegola2017knowledge} we honestly stated that our aim was not pushing the envelope on model accuracy, here we can --- also for the sake of honesty --- state that our aim was to squeeze the last ounce of AUC from the models. Still, such AUC-squeezing goal was tempered by aesthetic considerations: we did not want an overly complex, ugly solution, held together with shoestring and chewing gum. Added complexity had to bring proportional improvements in AUC, or we would prefer the simpler model.
   
\vspace{-0.1cm}	     
\subsection{Training data}
\label{sec:datasets}
    The freedom to use external sources makes the number of training samples a critical factor: deep models crave for data. We were prompted, in particular, by the much-publicized work of~\citet{esteva2017dermatologist}, which employed no less than 129,000 images, most of which unavailable to other researchers. 
    
    We collected several datasets to increase our training set. We restricted ourselves to publicly available (for free, or for a fee) reputable sources: 
      
    \begin{description}
        \item [ISIC 2017 Challenge] the official challenge dataset, with 2,000 dermoscopic images (374 melanomas, 254 seborrheic keratoses, and 1,372 benign nevi). 
        
        \item [ISIC Archive\footnote{The ISIC Archive: http://isdis.net/isic-project/}]
        with over 13,000 dermoscopic images.
                    
        \item [Interactive Atlas of Dermoscopy~{\normalfont\cite{argenziano2002dermoscopy}}] with 1,000+ clinical cases (270 melanomas, 49 seborrheic keratoses), each with at least two images: dermoscopic, and close-up clinical. 
        
        \item [Dermofit Image Library~{\normalfont\cite{ballerini2013color}}] with 1,300 images (76 melanomas, 257 seborrheic keratoses). 

        \item [IRMA Skin Lesion Dataset\footnote{IRMA datasets: http://ganymed.imib.rwth-aachen.de/irma/datasets/}]
        with 747 dermoscopic images (187 melanomas). 
        This dataset is unlisted, but available under special request, and the signing of a license agreement. 
    
        \item [PH2 Dataset~{\normalfont\cite{mendoncca2013ph}}] with 200 dermoscopic images (40 melanomas).
    \end{description}

\vspace{-0.1cm}	
    \subsubsection{Data debiasing (or rebiasing?)}
        Our first strategy, following our experience with deep learning, was to compose a training set as large as possible. Thus, we took all available images from all datasets, except those that could cause annotation clashes with the challenge (we excluded the images without diagnosis from the ISIC Archive, the `miscellaneous' class from the Atlas, the images marked as `benign' from IRMA, and the images marked as `atypical nevi' from PH2).
        
        We found, to our surprise, that such strategy was not optimal for melanoma, at least if measured by the validation AUC reported by the challenge website (\textbf{official validation AUCs}, hereafter). We found out a suspiciously large cluster of benign-lesion images at the ISIC Archive, all for 15-year old patients. Our validation numbers slightly improved after eliminating that cluster.
        
        We also found a distressing number of (near-)duplicates both inside and between the ISIC Challenge and the ISIC Archive datasets. The number was large enough to bias our \textbf{internal validation AUCs} --- i.e., the AUCs we got by evaluating in our internal validation dataset splits. Thus, we created a procedure to avoid train--test contamination, ensuring that all (near-)duplicates stayed in the same (training or validation)~split.
        
        We called \textbf{deploy} the dataset assembling all six sources, with the exclusions and deduplications explained above, resulting in 9,640 images.
        
        Nevertheless, best performance for melanoma on the official validation AUCs was, to our surprise, obtained with a dataset that assembled just ISIC Challenge, ISIC Archive, and Interactive Atlas (with the restrictions explained above, and some additional exclusions). We called \textbf{semi} this subset of deploy with 7544 images. For keratosis, the full dataset presented better results on the official validation AUCs.
        
        The results for melanoma raised difficult questions, since for the internal validation AUCs, the larger dataset was systematically better. Was the semi dataset just reflecting the official validation biases' (and if so, would those same biases be present in the official test set?) Or was semi actually better for melanoma? The Dermofit dataset --- present in deploy, but not in semi --- has a lot of carcinomas, which may confound the classification when added to the negative class. In the end we decided to keep the results from models trained both in deploy and in semi for the aggregate decision (Section~\ref{sec:metalearning}).
        
        Both deploy and semi datasets are precisely listed, image by image, in our \textbf{code repository}\footnote{Available soon at: https://github.com/learningtitans/isbi2017-part3}.

\vspace{-0.1cm}	
\subsection{Candidate Models}
    Our previous experience~\cite{menegola2017knowledge} showed that taking a model pre-trained on ImageNet, and fine-tuning it for skin lesion classification is a sound strategy to get good results. It also showed that better models for ImageNet (usually deeper and more expensive) tend to be also better for the newly fine-tuned~task.  
    
    We decided to concentrate our efforts in two models: ResNet-101~\cite{He2015} and Inception-v4~\cite{szegedy2016inceptionv4}. Both are state of the art, and are available in multiple frameworks, pre-trained for ImageNet with good results. The latter is important, because training from scratch on ImageNet can take weeks. We considered even larger models, like the gargantuan hybrid Inception--ResNet, but decided that its relative improvement on ImageNet was too small in comparison to its huge increase in memory~footprint.
    
    Part of the team conducted the experiments in Keras/Theano and part in Slim/Tensorflow, but at the end we merged all experiments into a single codebase in Slim/Tensorflow.
    
    We initially trained independent models for melanoma and for seborrheic keratosis, reflecting the structure of the challenge. However we switched to a single 3-class model as we realized it would be prohibitive to carry twice the training and evaluation for the entire experimental plan.

\vspace{-0.1cm}	
\subsection{Computational Resources}
    In order to perform a large number of trials, we attempted to secure as much computational horsepower as possible. For deep learning, that means large-memory CUDA-compatible GPUs.
    
    We intended to boost our installed capacity  --- currently RECOD has only half a dozen large-memory GPUs shared by over 60 students --- with rented servers. We found, however, no cloud service with the combination of affordability, availability to the public, and computing/memory capability (Amazon AWS: cheap and available; IBM SoftLayer: available and capable; Google Cloud: cheap and capable). 
    
    In the end we were saved by two fortuitous factors: (1) the LIP6/UPMC/Paris hosted Prof. Valle during most of the competition, and generously offered part of the needed resources; (2) it is Summer in the Southern Hemisphere, reducing competition for RECOD GPUs during the big vacations.
    
\vspace{-0.1cm}	
\subsection{Experimental \st{Design} Tactics}
    We attempted to conciliate our inherent scientific drive, with the sportive --- but non-scientific --- ambition to win the challenge, or at least to maximize our chances of doing so. We decided to proceed with a mix of pragmatism and rigor. On one hand, there simply would not be enough time for significance tests or ANOVAs; we would have to postpone those. On the other hand, we still wanted to make sound decisions along the way, and to be able to reproduce any results later on.
    
    The team used Slack (a chat service for professionals) as main channel for communication. We coordinated the tasks with Google Docs, and shared the results of each intermediate experiment with Google Sheets. We used code version control (with git) to facilitate future reproduction of intermediate steps.
    
    We established an initial agenda of hypotheses to validate. Omitting a few speculative ideas we did not have time to touch, those were: 
    
    \begin{enumerate}
        \item Compare the baseline VGG-16 network to the deeper ResNet-101 or Inception-v4;
        \item Compare standard-resolution images (224 for VGG and ResNet) to double-resolution images;
        \item Contrast different strategies of class- and sample- weighting during training;
        \item Compare normal training schedule with some form of curriculum-learning;
        \item Contrast different regimens of training and test augmentation;
        \item Measure the impact of adding SVM as a final decision~layer;
        \item Attempt to use the patient data (age and sex) on classification;
        \item Attempt different model optimizers;
        \item Add different types of per-sample normalization;
        \item Add a final meta-decision based upon multiple models (ensemble, stacking, etc.)
    \end{enumerate}

    Our normal procedure would be to attempt an (incomplete) factorial design, at least for the factors where we expected cross-effects (e.g., depth $\times$ resolution $\times$ weighting $\times$ scheduling $\times$ augmentation). For the competition, however, there was no time for such level of rigor. We tested the hypotheses more or less sequentially, revisiting only those that seemed too surprising, and crossing only the effects for which we had a very strong expectation for interactions.

    \noindent\textbf{Serendipity:} The tests with the semi-dataset (Section~\ref{sec:datasets}) were not programmed from the beginning, they were a happenstance due to the delay in obtaining the complete dataset. When extra data became available, we noticed a sharp drop in the official validation AUC for melanoma (but not for keratosis), and decided to investigate.

\vspace{-0.1cm}	
\subsection{Failures to Thrive}

    Most of our attempts resulted in little to none improvement. We were not very diligent, however, in pursuing any factor whose effect size seemed small, and we performed no significance nor equivalence tests. \emph{Caveat lector} before any definitive conclusion about the ``uselessness'' of the factors below.

    We sort the list placing first the biggest disappointments/surprises --- the factors we most expected to improve the results but did not:
    \begin{enumerate}
        \item Image resolution: we tried both amending VGG-16 to accept larger inputs, and amending the augmentation procedure of Inception-v4 to accept larger images pre-cropping (but keeping the network input itself unchanged). Neither attempt improved the results.
        \item We attempted several class- and sample- weighting schemes, both to compensate the unbalancing of the classes, and the reliability of the annotations. In one case we attributed weights inversely proportional to the frequency of the classes; in another case we combined those with weights that went from 1 to 3 ranging from ``unknown follow-up''/``no follow-up'' until ``confirmed by histopathology'' (we attributed 5 to the official dataset to give it extra weight). The more complex the weighting scheme, the worse the AUCs --- no weighting was the best weighting. 
        \item Validation and early stopping: we tried two ways to perform early stopping: first, when our internal validation AUC started to decrease, and second (more aggressive) when it refused to increase. With a single exception, there was no impact in the results. We could not afford the \emph{very} long fine-tunings (several weeks) recommended for Inception in some applications\footnote{https://github.com/tensorflow/models/tree/3be9ece9574d7bac07704e$\dlsh$ \\43705741d9af1de1e6/im2txt\#fine-tune-the-inception-v3-model}. 
        It is possible that in those super-long training scenarios early stopping with validation becomes important.
        \item Patient data: we attempted different encodings for incorporating the patient data (age and sex) into the features, inserting them in the transition between the deep model and the SVM decision layer. The results were inconsistent, sometimes improving and sometimes worsening the results.
        \item Curriculum learning: curriculum learning consists in careful scheduling of the training samples in order to present a ``curriculum'' of learning steps to the algorithm, instead of learning the samples at random (e.g., learning the easy cases first). The Interactive Atlas' samples are annotated with a level of diagnosis difficulty (from a human point of view), allowing such scheme. We attempted a three-step schedule (starting with Atlas' easy images, proceeding to Atlas' easy and moderate images, and finalizing with all images). The results were worse than simply training with all images at once. 
        \item Segmentation information: we intended to incorporate the segmentation model learned in Part 1 in our classifier, and designed several more-or-less complex ways to do it. Unfortunately, due to the time limitations, the only experiments we performed were too quick-n-dirty to work. We still believe that properly encoded segmentation information could help.
    \end{enumerate}

\begin{figure*}[t!]
    \centering
    \subfigure{\includegraphics[width=.315\textwidth]{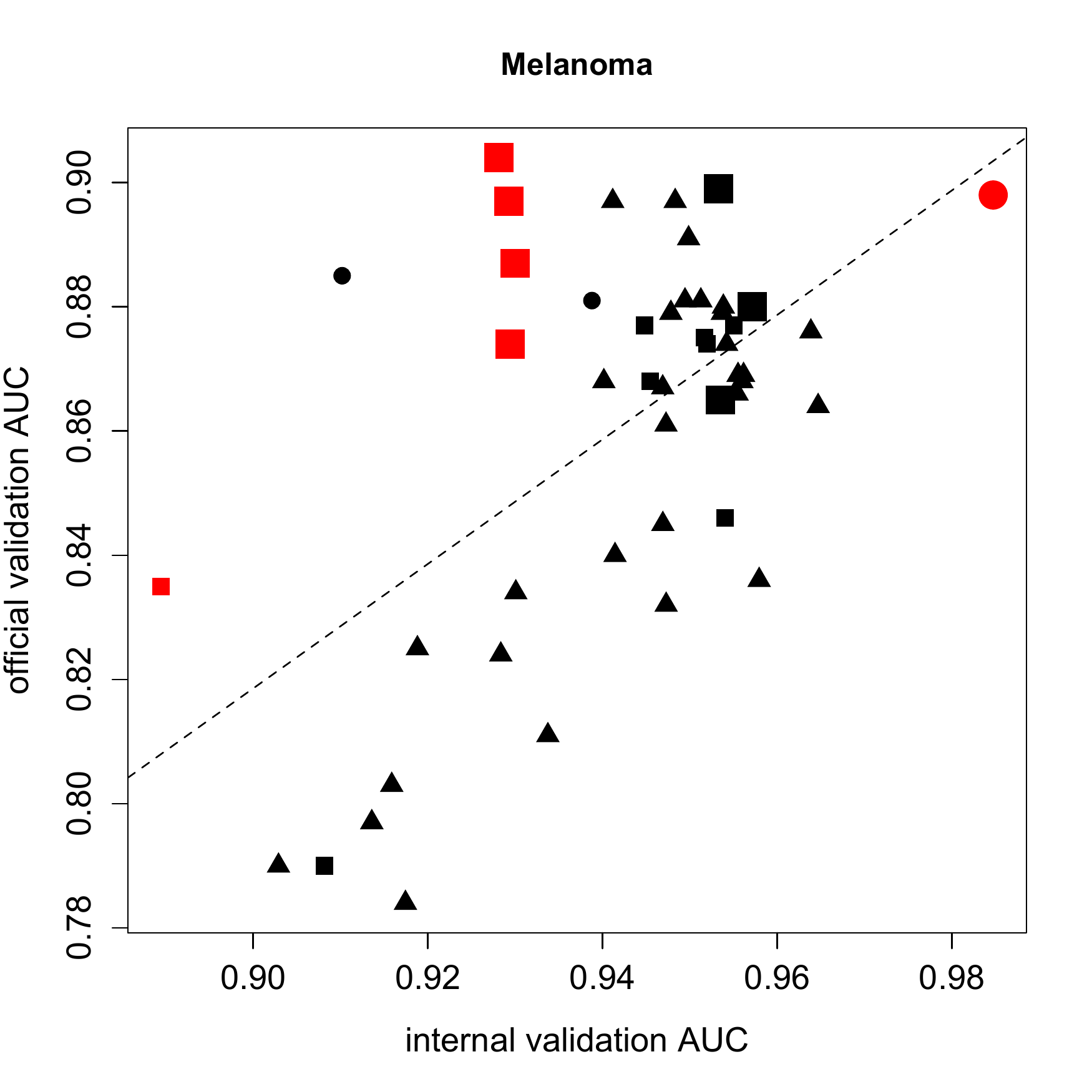}} 
    \subfigure{\includegraphics[width=.315\textwidth]{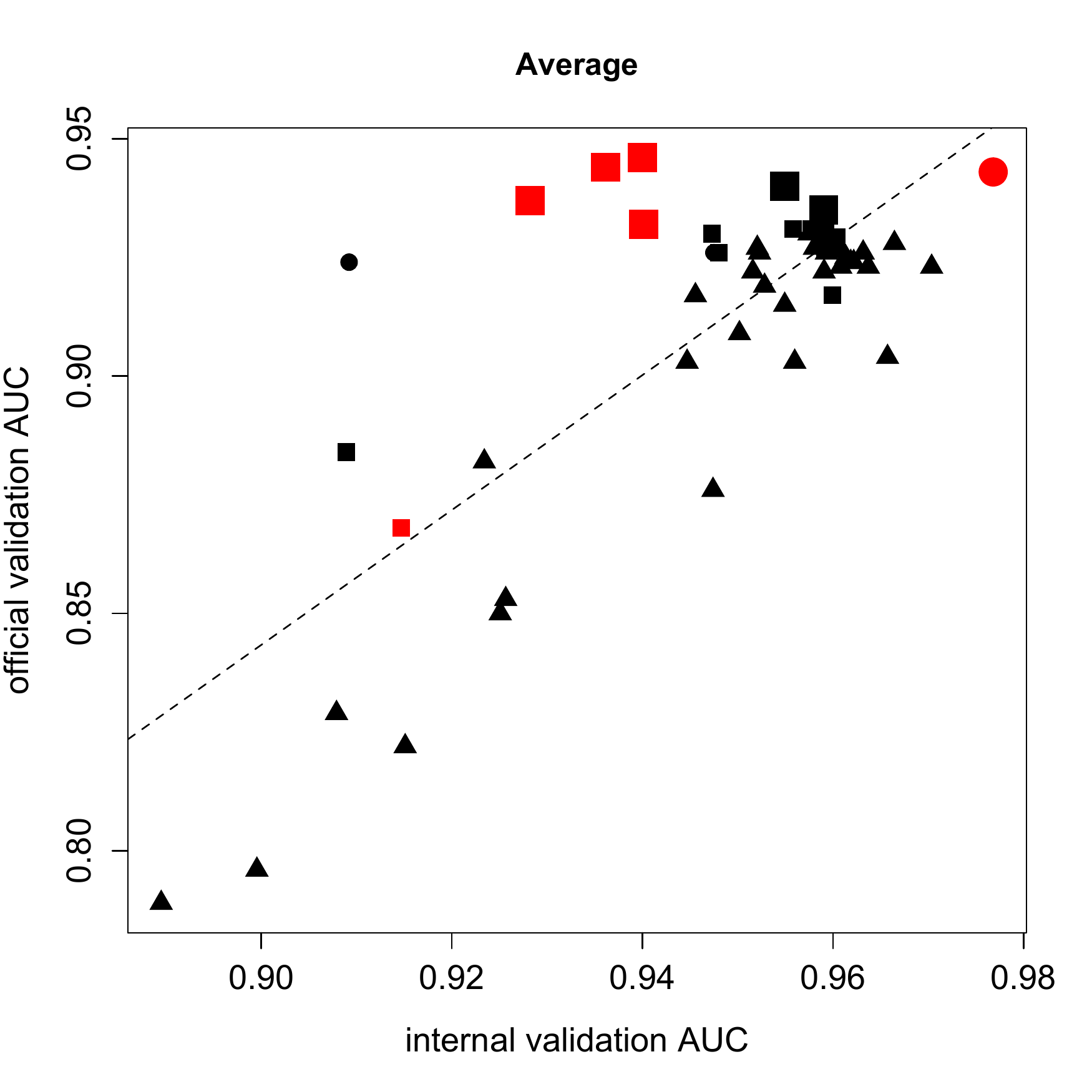}} 
    \subfigure{\includegraphics[width=.315\textwidth]{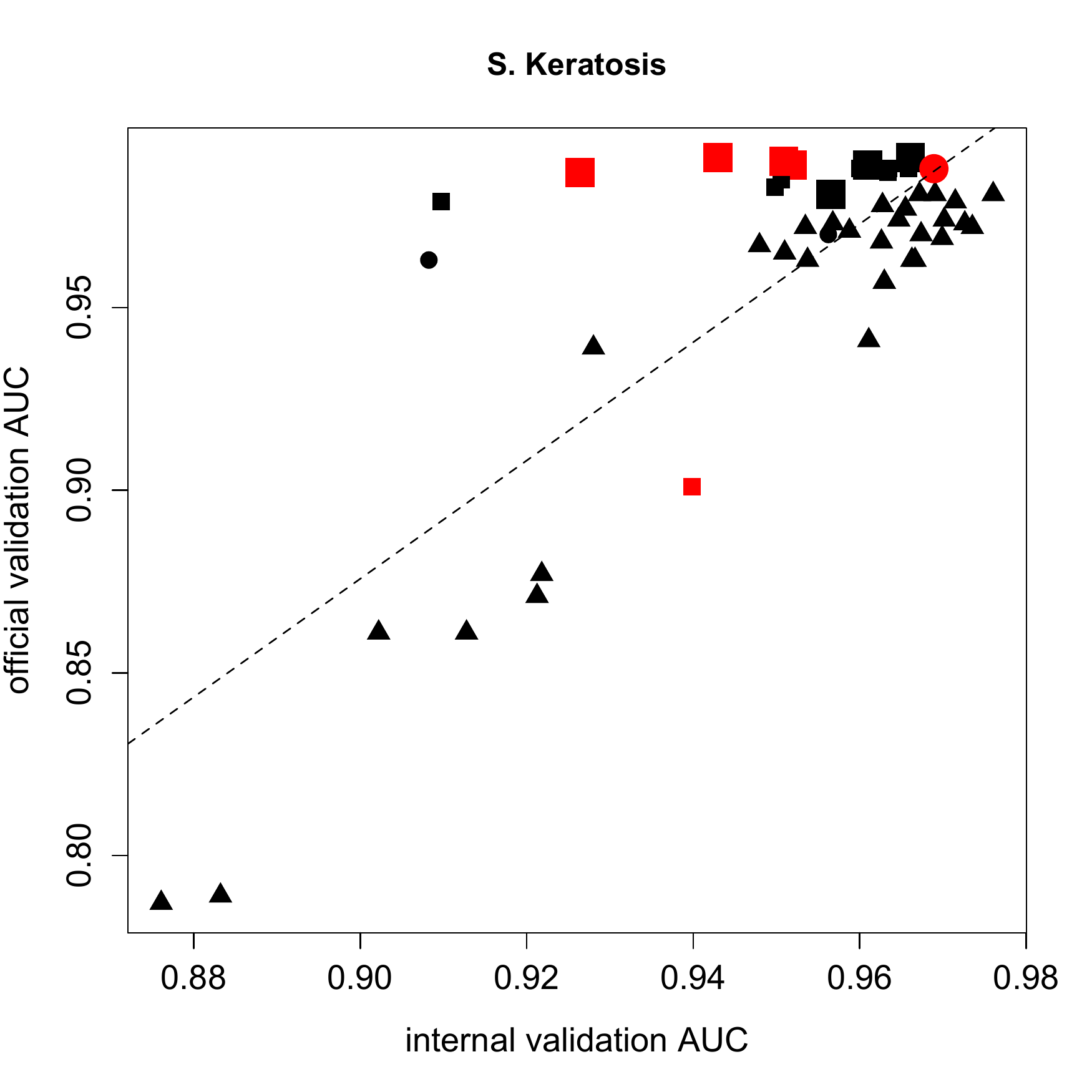}} 
    \caption{A visual panorama of our experiments. The circles are ResNet-101 Models, the triangles and squares are Inception-101 models (without and with per-image normalization respectively). Black and red indicates training in the deploy and semi datasets respectively. Large symbols indicate the models chosen to compose the stacking in the final submission. The dashed line is the regression between the internal and the official AUCs. The models are the same in the three graphs, but the metrics change from Melanoma, Average, and Seborrheic Keratosis AUC from left tor right. Those were only a subset of the experiments, 48 out of more than a hundred models we attempted.}
    \label{fig:plots}
\end{figure*}

\vspace{-0.1cm}	
\subsection{Success Factors}

    If most attempts disappointed, some definitely were valuable, as measured by both the internal and official validation AUCs. We have not, for the moment, a factorial analysis to quantify the contributions, but we sort the list placing first the factors we believe helped the most:
    \begin{enumerate}
        \item Models + data: the mere transition to deeper models helped, but not by very much. It was the combination of deeper models \emph{and} larger datasets that boosted the numbers. Our smaller semi dataset is already several times over the size of the official challenge training set.
        \item Data augmentation: from experience, we knew that training with data augmentation is critical (i.e., applying random transformations: croppings, flippings, etc. on the images before using them in the network) and made all attempts with it. Train augmentation is not set to a fixed number of transformations: as long as the training persists, images are sampled from the training set, and random transformations are applied to them. We found out that \emph{test} augmentation is critical as well: applying random transformations to the test sample, submitting those transformed samples to the network, and then pooling the results. When we employed an SVM decision layer after the network, augmentation was again fruitful, and when we stacked several models with a meta-learning SVM, augmentation was yet again important. We attempted several schemes for pooling, but a simple average pooling worked best in all cases.
        \item Per-image normalization: on Inception, normalizing the inputs to the network by subtracting the average image pixel improved results considerably. Surprisingly, going one step further and dividing the pixels by the standard deviation gave worse results than no per-image normalization at all.  We did not have time to test this factor on ResNet.
        \item Stacking models and meta-learning: fusing the decision of several models gave, almost always, better results than just using the single best model, even when using simple schemes, like averaging the probabilities among the models for a given sample. However, a meta-learning scheme, using an additional SVM layer to learn the decision from the probabilities output by the models, gave the best results on the official validation AUC.
    \end{enumerate}

\vspace{-0.1cm}	
\subsection{Assembling the Final Submission}
\label{sec:metalearning}

    Figure~\ref{fig:plots} shows a subset of 48 out of more than a hundred models we evaluated (most experiments were too quick-and-dirty to allow inclusion in the plot). From the beginning we noticed that the correlation between our internal validation AUCs and the official validation AUCs was far from perfect. In the plots shown, from left to right, the correlations are R=0.58, R=0.77, and R=0.79. The correlation was particularly bad for melanoma. That posed a challenge of choosing who to trust: the official or the internal validation AUC. In the end we chose to trust both (or neither), and included models that showed good performance in the two axes. 
    
    Another difficulty was that the best models for melanoma were not necessarily those for keratosis and vice-versa. We considered selecting different models for the different tasks, but in the end we decided to pick the same set of models for both tasks and hope the meta-learning layer would do the adjustments.
    
    The meta-learning consisted in, for each sample, concatenating the decisions of each chosen model and using this as feature vector for two binary SVMs (melanoma-vs-all, keratosis-vs-all). Those SVMs were trained using our internal validation set --- thus we were prevented from evaluating them using the internal validation AUC. However this scheme attained the best official validation AUC.
    
    We attempted several small variations for the meta-learning; the most successful employed an aggressive augmentation scheme: each training sample (from our internal validation set) was evaluated thrice by each model, allowing to create a large number of combined replicas for training. The same procedure was applied for testing (on the samples from the official validation and official test sets). In both cases, we employed average pooling to combine the replicas.
    
    \textbf{The submitted test run} as well as our last official validation run were, thus, the result from a meta-model that assembled seven base models: three based on Inception trained on deploy; three based on Inception trained on semi; and one based on ResNet trained on semi. The results of those component models were stacked in a meta-learning layer based on an SVM trained on the validation set of deploy.
    
	

\vspace{-0.2cm}
\section*{Acknowledgements}
    {\small
        A. Menegola is funded by CNPq; S. Avila is funded by PNPD/CAPES; M. Fornaciali and E. Valle are partially funded by Google Research Awards for Latin America 2016; E. Valle is also partially funded by a CNPq PQ-2 grant (311486/2014-2). RECOD Lab. is partially supported by diverse projects and grants from FAPESP, CNPq, and CAPES. We gratefully acknowledge the donation of a Tesla K40 GPU by NVIDIA Corporation, used in this work.
        Prof. Eduardo Valle was hosted by LIP6/UPMC/Paris in January/February 2017, and was kindly offered access to its GPU infrastructure, extensively used in this work, by Prof. Matthieu Cord. We are grateful to RECOD and LIP6 members --- and in particular to George Gondim, Prof. Matthieu Cord, Micael Carvalho, Pedro Tabacof, Ramon Oliveira and Ramon Pires --- for scientific and technical insights on machine learning. We thank Prof. Flávia V. Bittencourt for the insights on dermoscopy and lesion diagnosis. We thank Prof. M. Emre Celebi for kindly providing the machine-readable metadata of The Interactive Atlas of Dermoscopy. 
    }
    
\bibliographystyle{IEEEtranN}
\bibliography{references}

\begin{thebibliography}{13}
\providecommand{\natexlab}[1]{#1}
\providecommand{\url}[1]{#1}
\csname url@samestyle\endcsname
\providecommand{\newblock}{\relax}
\providecommand{\bibinfo}[2]{#2}
\providecommand{\BIBentrySTDinterwordspacing}{\spaceskip=0pt\relax}
\providecommand{\BIBentryALTinterwordstretchfactor}{4}
\providecommand{\BIBentryALTinterwordspacing}{\spaceskip=\fontdimen2\font plus
\BIBentryALTinterwordstretchfactor\fontdimen3\font minus
  \fontdimen4\font\relax}
\providecommand{\BIBforeignlanguage}[2]{{%
\expandafter\ifx\csname l@#1\endcsname\relax
\typeout{** WARNING: IEEEtranN.bst: No hyphenation pattern has been}%
\typeout{** loaded for the language `#1'. Using the pattern for}%
\typeout{** the default language instead.}%
\else
\language=\csname l@#1\endcsname
\fi
#2}}
\providecommand{\BIBdecl}{\relax}
\BIBdecl

\bibitem[Fornaciali et~al.(2014)Fornaciali, Avila, Carvalho, and
  Valle]{fornaciali2014statistical}
M.~Fornaciali, S.~Avila, M.~Carvalho, and E.~Valle, ``Statistical learning
  approach for robust melanoma screening,'' in \emph{{SIBGRAPI}}, 2014, pp.
  319--326.

\bibitem[Carvalho(2015)]{carvalho2015}
M.~Carvalho, ``Transfer schemes for deep learning in image classification,''
  Master's thesis, University of Campinas, 2015.

\bibitem[Fornaciali et~al.(2016)Fornaciali, Carvalho, Bittencourt, Avila, and
  Valle]{fornaciali2016towards}
M.~Fornaciali, M.~Carvalho, F.~V. Bittencourt, S.~Avila, and E.~Valle,
  ``Towards automated melanoma screening: Proper computer vision \& reliable
  results,'' \emph{arXiv:1604.04024}, 2016.

\bibitem[Menegola et~al.(2017)Menegola, Fornaciali, Pires, Bittencourt, Avila,
  and Valle]{menegola2017knowledge}
A.~Menegola, M.~Fornaciali, R.~Pires, F.~V. Bittencourt, S.~Avila, and
  E.~Valle, ``Knowledge transfer for melanoma screening with deep learning,''
  in \emph{IEEE ISBI}, 2017.

\bibitem[Ronneberger et~al.(2015)Ronneberger, Fischer, and
  Brox]{RonnebergerFB15}
O.~Ronneberger, P.~Fischer, and T.~Brox, ``U-net: Convolutional networks for
  biomedical image segmentation,'' \emph{arXiv:1505.04597}, 2015.

\bibitem[Simonyan and Zisserman(2014)]{DBLP:journals/corr/SimonyanZ14a}
K.~Simonyan and A.~Zisserman, ``Very deep convolutional networks for
  large-scale image recognition,'' \emph{arXiv:1409.1556}, 2014.

\bibitem[Codella et~al.(2015)Codella, Cai, Abedini, Garnavi, Halpern, and
  Smith]{codella2015deep}
N.~Codella, J.~Cai, M.~Abedini, R.~Garnavi, A.~Halpern, and J.~R. Smith, ``Deep
  learning, sparse coding, and svm for melanoma recognition in dermoscopy
  images,'' in \emph{MLMI}, 2015, pp. 118--126.

\bibitem[Esteva et~al.(2017)Esteva, Kuprel, Novoa, Ko, Swetter, Blau, and
  Thrun]{esteva2017dermatologist}
A.~Esteva, B.~Kuprel, R.~A. Novoa, J.~Ko, S.~M. Swetter, H.~M. Blau, and
  S.~Thrun, ``Dermatologist-level classification of skin cancer with deep
  neural networks,'' \emph{Nature}, vol. 542, no. 7639, pp. 115--118, 2017.

\bibitem[Argenziano et~al.(2002)Argenziano, Soyer, De~Giorgi, Piccolo, Carli,
  Delfino, et~al.]{argenziano2002dermoscopy}
G.~Argenziano, H.~P. Soyer, V.~De~Giorgi, D.~Piccolo, P.~Carli, M.~Delfino
  \emph{et~al.}, ``Dermoscopy: a tutorial,'' \emph{EDRA, Medical Publishing \&
  New Media}, 2002.

\bibitem[Ballerini et~al.(2013)Ballerini, Fisher, Aldridge, and
  Rees]{ballerini2013color}
L.~Ballerini, R.~B. Fisher, B.~Aldridge, and J.~Rees, ``A color and texture
  based hierarchical k-nn approach to the classification of non-melanoma skin
  lesions,'' in \emph{CMIA}, 2013, pp. 63--86.

\bibitem[Mendon{\c{c}}a et~al.(2013)Mendon{\c{c}}a, Ferreira, Marques, Marcal,
  and Rozeira]{mendoncca2013ph}
T.~Mendon{\c{c}}a, P.~M. Ferreira, J.~S. Marques, A.~R. Marcal, and J.~Rozeira,
  ``{PH 2} — a dermoscopic image database for research and benchmarking,'' in
  \emph{IEEE EMBC}, 2013, pp. 5437--5440.

\bibitem[He et~al.(2015)He, Zhang, Ren, and Sun]{He2015}
K.~He, X.~Zhang, S.~Ren, and J.~Sun, ``Deep residual learning for image
  recognition,'' \emph{arXiv:1512.03385}, 2015.

\bibitem[Szegedy et~al.(2016)Szegedy, Ioffe, and
  Vanhoucke]{szegedy2016inceptionv4}
C.~Szegedy, S.~Ioffe, and V.~Vanhoucke, ``Inception-v4, inception-resnet and
  the impact of residual connections on learning,'' \emph{arXiv:1602.07261},
  2016.

\end{thebibliography}
    
\end{document}